\documentclass{article}

\usepackage{microtype}
\usepackage{amsmath,amssymb}
\usepackage{graphicx}
\usepackage{subfigure}
\usepackage{booktabs} 

\usepackage[accepted,nohyperref]{icml2019}

\icmltitlerunning{DeepClimGAN: A High-Resolution Climate Data Generator}

\begin{document}

\twocolumn[
\icmltitle{DeepClimGAN: A High-Resolution Climate Data Generator}




\begin{icmlauthorlist}
\icmlauthor{Alexandra Puchko}{wwu}
\icmlauthor{Robert Link}{jgcri}
\icmlauthor{Brian Hutchinson}{wwu,pnnl}
\icmlauthor{Ben Kravitz}{indiana}
\icmlauthor{Abigail Snyder}{jgcri}
\end{icmlauthorlist}

\icmlaffiliation{wwu}{Computer Science Department, Western Washington University, Bellingham, WA}
\icmlaffiliation{jgcri}{Joint Global Change Research Institute, Pacific Northwest National Laboratory, College Park, MD}
\icmlaffiliation{pnnl}{Computing and Analytics Division, Pacific Northwest National Laboratory, Seattle, WA}
\icmlaffiliation{indiana}{Earth and Atmospheric Sciences Department, Indiana University, Bloomington, IN}

\icmlcorrespondingauthor{Brian Hutchinson}{Brian.Hutchinson@wwu.edu}

\icmlkeywords{Machine Learning, ICML}

\vskip 0.3in
]



\printAffiliationsAndNotice{}  

\section{Introduction}

Earth system models (ESMs), which simulate the physics and
chemistry of the global atmosphere, land, and ocean, are often used to generate future projections of climate change scenarios. These models
are far too computationally intensive to run repeatedly, but limited sets of runs are insufficient for some important applications, like adequately sampling distribution tails to characterize extreme events. As a compromise, emulators are substantially less expensive but may not have all of the complexity of an ESM. Here we demonstrate the use of a conditional generative adversarial network (GAN)
to act as an ESM emulator. In doing so, we gain the ability to produce daily weather data that is consistent with what ESM might output over any chosen scenario. In particular, the GAN is aimed at representing a joint probability distribution over space, time, and climate variables, enabling the study of correlated extreme events, such as floods, droughts, or heatwaves.

The use of neural networks in weather forecasting predates the deep learning
boom \cite{hall_brooks, Koizumi}.  \citet{2015arXiv150604214S}
introduced convolutional LSTMs for the task of precipitation nowcasting; we plan to incorporate
elements from their architecture into our design.
\citet{gmd-11-3999-2018} and
\citet{Schneider} both present important considerations and challenges in
weather and climate modeling with machine learning.
\citet{2019arXiv190506841W} present a strategy for incorporating constraints
into GANs that are applicable to our task. \citet{rasp} and
\citet{Brenowitz} both demonstrate that deep
learning can be used to accurate model subgrid processes in climate tasks.
Neither use GANs, but \citet{rasp} expresses optimism about their potential.
Both \citet{acp} and \citet{gmd-12-1791-2019}
tackle efficient surrogate modeling, and find deep learning to be
effective.

\section{High-Resolution Climate Data Generation} \label{sec:approach}

\subsection{The DeepClimGAN}

Generative Adversarial Networks (GANs) \cite{2014arXiv1406.2661G} have been
rapidly and widely adopted for the generation of realistic images. They
leverage two competing architectures: the generator and the discriminator.
The networks are trained jointly in a minimax fashion, ideally reaching an
equilibrium in which samples from the two distributions are
indistinguishable and the discriminator cannot exceed 50\% accuracy.
GANs have had widespread success in image and video applications,
making them a promising choice for generating gridded climate data.

\emph{DeepClimGAN} is a conditional GAN, capable of producing a spatio-temporal forecast, generating samples $y \in \mathbb{R}^{H \times W \times T \times V}$, where the spatial dimensions are $H$ and $W$, the temporal dimension is $T$, and there are $V$ real-valued climate variables predicted at each location and time. 
For simplicity, we set $T=32$ for a convenient ``month-like'' forecast period. The $V=7$ climate variables we use in this work are: $\lbrace$ min, avg, max $\rbrace$ temperature, $\lbrace$ min, avg, max $\rbrace$ relative humidity, and precipitation; we chose these variables because those are the variables required by various impact models, including hydrology, agriculture and health (e.g. heat index).
A high-resolution generator for these variables would enable new
insights into climate impacts and risks faced by human systems.

Each sample is conditioned on some context, $c$. This conditioning information should capture the initial state from the
start of the forecast and the constraints we want our forecast to observe. Therefore, we assume that $c$ consists of two components:
1) ``monthly'' context $c_1 \in
\mathbb{R}^{H \times W \times 2}$, containing the average precipitation and
temperature for the month (length $T$ period), and 2) recent context $c_2
\in \mathbb{R}^{H \times W \times KV}$, containing all $V$ climate variables
for the $K$ days immediately preceding the month, stacked.
The the $c_1$ information allows
us to specify the type of scenario (e.g. high or low warming) we wish
to generate, while the $c_2$ information allows us to ensure continuity between months.

\begin{figure*}[t]
  \centering
  \includegraphics[width=0.8\textwidth]{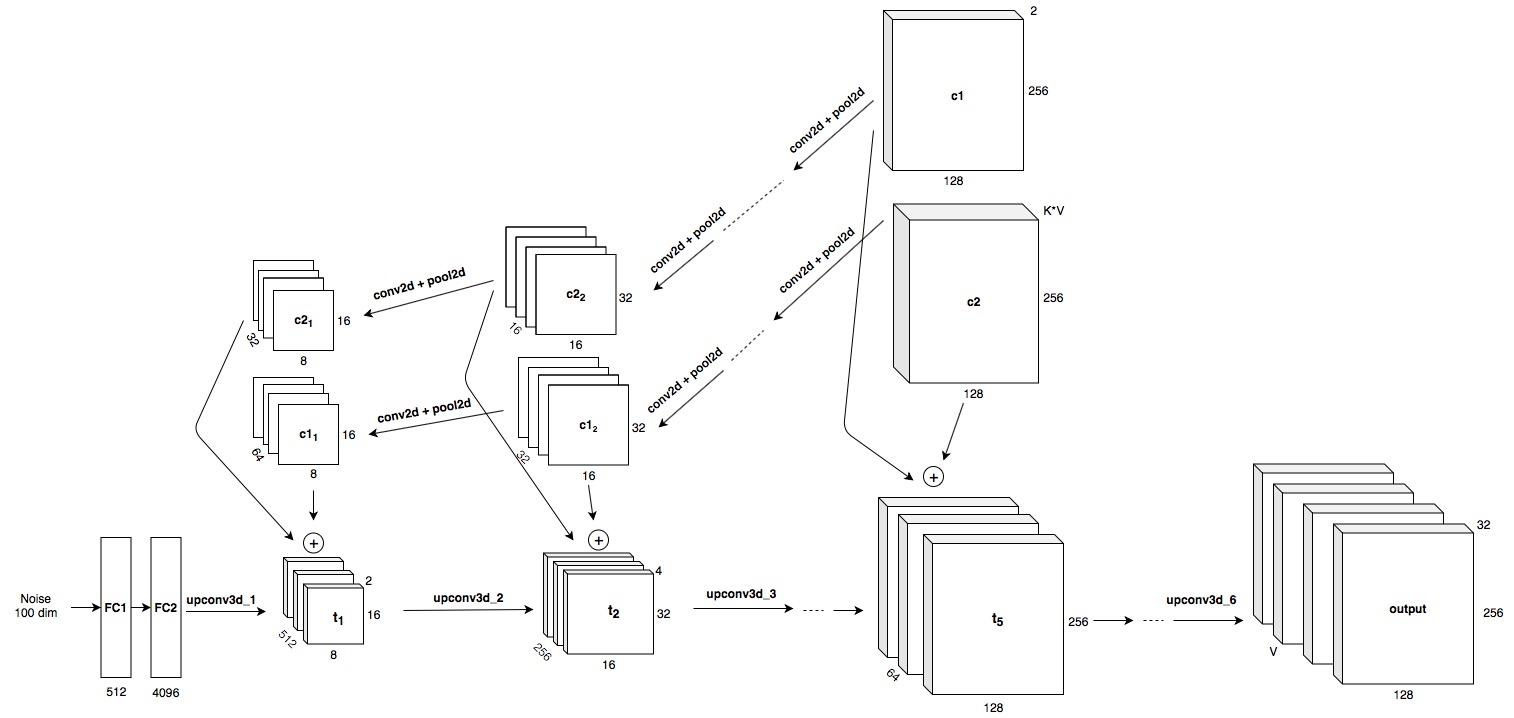}
  \caption{Generator.}
  \label{fig:generator}
\end{figure*}

\begin{figure}
  \centering
  \includegraphics[width=8cm]{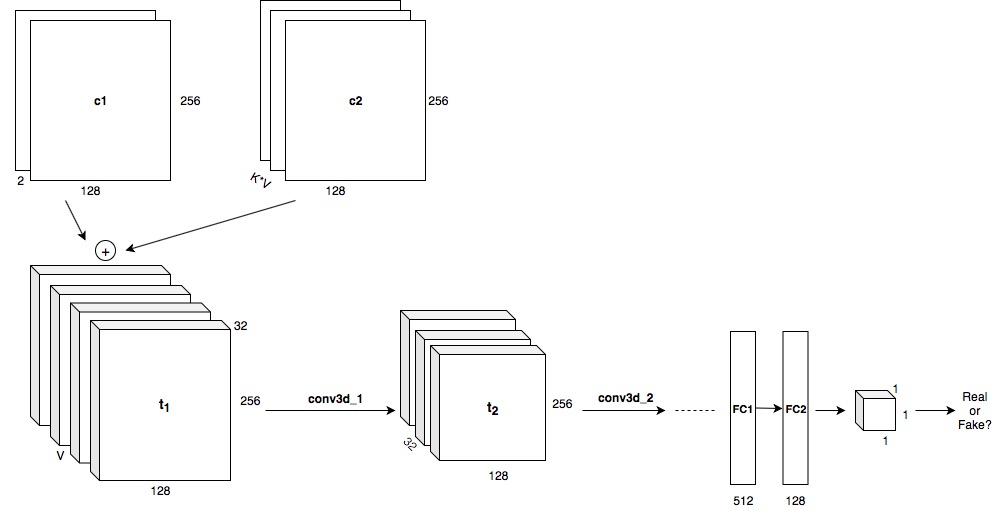}
  \caption{Discriminator.}
  \label{fig:discriminator}
\end{figure}

The generator and discriminator architectures, including conditioning
subarchitectures, are shown in Figures~\ref{fig:generator} and
\ref{fig:discriminator}, respectively, and described below. Both architectures were inspired by the work of \citet{videogan} on video generation with GANs.

\subsubsection*{Generator} \label{generator}

As shown in the bottom track of Fig.~\ref{fig:generator}, the base generator projects noise $z \in \mathbb{R}^{100}$ to $\mathbb{R}^{4096}$ via two fully connected layers; it is then reshaped and fed through a series of six 3d up (transposed) convolutions.  After all but the last layer, we apply batch
normalization \cite{bnorm} followed by a ReLU activation function. For
the output layer, we apply ReLU only to the precipitation channel to,
avoid predicting negative precipitation, while still allowing
precipitation-free days.
Additionally, we pass the mean, maximum and minimum relative humidity through a logistic sigmoid to constrain them to be in $(0,1)$.

As described above, we incorporate two forms of context:
low resolution monthly totals $c_{1}$, and 
a high resolution initial condition $c_{2}$.
These contexts are injected at each convolutional layer into the base generator.
The original contexts $c_1$ and $c_2$ are spatially downsampled through
a series of 2d convolutional and $2\times 2$ maxpool layers.
We replicate the 2d contexts across the time dimension and then append along the channel dimension so that it serves as additional input to each 3d up convolution.
The end result, in the lower right of Fig.~\ref{fig:generator}, is our spatio-temporal forecast for all $V$ climate variables.

\subsubsection*{Discriminator} \label{discriminator}
The discriminator, shown in Fig.~\ref{fig:discriminator}, consists of four 3d convolution layers followed by two fully connected layers.
Each layer is followed by batch normalization \cite{bnorm} and leaky
ReLU \cite{leakyrelu}. The output activation is logistic sigmoid,
yielding a scalar probability of the data being real (i.e., instead of
generated).
Conditioning is accomplished by appending the context along the channel dimension of the input.

\subsubsection*{Training} \label{training}
We train the model with alternating updates for generator and discriminator.
Our true examples are 32 day periods with randomly sampled start days.
The high resolution context ($c_2$) is the five days prior to the sampled month. The low resolution context is produced by averaging precipitation and temperature over the 32 day period.
The model is implemented in PyTorch, based on a model described in \citet{videogan}.
We use Adam \cite{kingma2014adam} to optimize the model. The learning rate is fixed at 0.0002 with a momentum of 0.5 and a batch size of 32. We initialize all weights with zero mean Gaussian noise with standard deviation 0.02. To improve training, we have explored 1) adding Gaussian noise to the real and generated data before feeding to the discriminator, 2) adding an experience replay mechanism, in which previously generated samples are stored and also fed to the discriminator, and 3) pretraining the generator to produce similar marginal temperature and precipitation statistics as ground truth data.

\subsection{Generation with the DeepClimGAN}

We generate high-resolution weather arbitrarily far into the future as follows:
\begin{enumerate}
\item \setlength{\parskip}{-3pt} \label{step1} Run a low-resolution
    ESM emulator to produce monthly contexts ($c_1s$) for as many future months as desired.
\item \label{stepN} Sample month 1's weather from the generator, conditioned on $K$ days of ground truth $c_2$ and the first month's low resolution context $c_1$.
\item For $i=2 \ldots N$: take the last $K$ days of the previous month's generated data as the new $c_2$, input the corresponding $c_1$ from the low-resolution emulator and sample month $i$'s weather from the generator.
\end{enumerate}

The rationale for using the initial state context is that it allows us to maintain continuity: when we chain
several months of generated data together, conditioning on the last few days
of the previous month allows us to avoid having statistical artifacts at the
month boundary.

\section{Experiments and Initial Findings}
\subsection{Data} \label{data}
We used daily resolution CMIP5 (``Coupled Model Intercomparison Project, Phase 5'') \cite{cmip5} archival data for the MIROC5 model
\cite{watanabe2010improved} under different greenhouse gas emission scenarios. The scenarios provided by the model span: pre-industrial (idealized, constant 1860 conditions, run for 200 years), historical (1861-2005), future (2006-2099), and extended future (2100-2299). 
Each scenario is represented in several realizations (simulations with different initial conditions to capture a range of internal variability).
Details of the source data are provided in Table~\ref{tab:data} in Appendix~B.
DeepClimGAN is agnostic to the model it emulates: swapping the training data should result in a generator that can approximate the desired data distribution. We selected 1680 years of data for training. A daily map for each of the climate variables is represented in 128 $\times$ 256 spatial resolution. We applied $\log(1+x)$-normalization for precipitation, mapped relative humidity into $(0,1)$ and standardized the temperature variables.

\subsection{Initial Findings} \label{findings}
\begin{figure}[tb]
    \centering
    \includegraphics[width=0.30\textwidth]{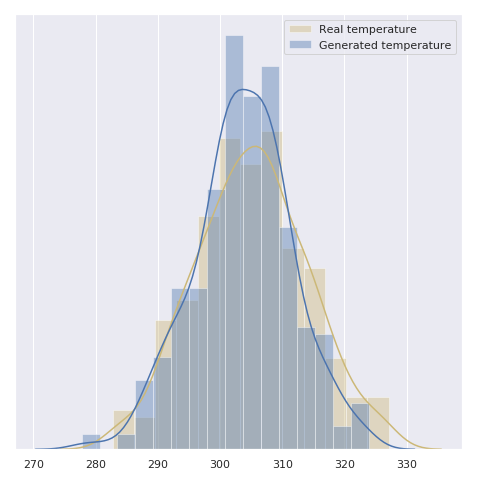}
    \caption{Histogram of real vs generated daily avg. temperature.}
    \label{fig:tempresults}
\end{figure}

\begin{figure}[tb]
    \centering
    \includegraphics[width=0.7\columnwidth]{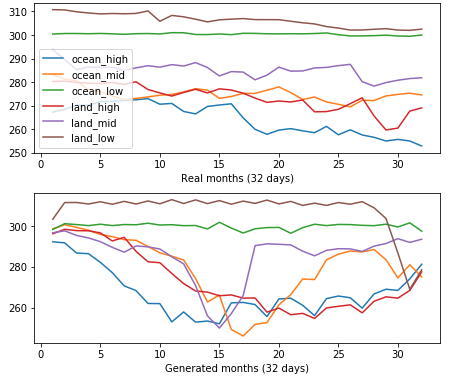}
    \caption{Real (top) and generated (bottom) daily temperature (K).}
    \label{fig:generatedtemp}
\end{figure}

Figures~\ref{fig:tempresults} and \ref{fig:generatedtemp} show the results of initial experiments, with a model that includes pretraining as described in Sec. \ref{sec:approach}. Fig.~\ref{fig:tempresults} shows histograms of temperature over 320 randomly sampled 32-day periods, contrasting ground truth with generator output. 
Fig.~\ref{fig:generatedtemp} shows real and generated daily temperatures for one 32 day period, for six spatial locations, each belonging to a distinct region.
Both figures suggest that the generator is capable of generating fairly realistic daily average temperature values. 

\section{Future Work}
Our initial findings are encouraging, but there remains a great deal of important work for this on-going project.
First, we must continue to develop and refine evaluation metrics for this task. While evaluating GANs is notoriously challenging, our use case opens the door to some promising approaches. In addition to the ideas discussed in Appendix~A, we will explore ways to quantify the generation quality for downstream applications (e.g. characterizing extreme events).
Second, extensive experimentation is needed in order to assess the model's ability to model the distributions induced by different climate emulators and climate change scenarios.
Third, there are other promising architectures to be considered for the discriminator and generator, including those based on convolutional LSTMs.
Fourth, there is rich literature of techniques we plan to draw from to improve the generative quality of our GANs (e.g. label smoothing, historical averaging, modified cost functions).
Finally, we plan to generate datasets and disseminate both the data and DeepClimGAN tool to facilitate climate change research.

\bibliography{nips2019_icml}

\begin{thebibliography}{19}
\providecommand{\natexlab}[1]{#1}
\providecommand{\url}[1]{\texttt{#1}}
\expandafter\ifx\csname urlstyle\endcsname\relax
  \providecommand{\doi}[1]{doi: #1}\else
  \providecommand{\doi}{doi: \begingroup \urlstyle{rm}\Url}\fi

\bibitem[Brenowitz \& Bretherton(2018)Brenowitz and Bretherton]{Brenowitz}
Brenowitz, N. and Bretherton, C.
\newblock Prognostic validation of a neural network unified physics
  parameterization.
\newblock 05 2018.
\newblock \doi{10.31223/osf.io/eu3ax}.

\bibitem[Dueben \& Bauer(2018)Dueben and Bauer]{gmd-11-3999-2018}
Dueben, P.~D. and Bauer, P.
\newblock Challenges and design choices for global weather and climate models
  based on machine learning.
\newblock \emph{Geoscientific Model Development}, 11\penalty0 (10):\penalty0
  3999--4009, 2018.
\newblock \doi{10.5194/gmd-11-3999-2018}.

\bibitem[{Goodfellow} et~al.(2014){Goodfellow}, {Pouget-Abadie}, {Mirza}, {Xu},
  {Warde-Farley}, {Ozair}, {Courville}, and {Bengio}]{2014arXiv1406.2661G}
{Goodfellow}, I.~J., {Pouget-Abadie}, J., {Mirza}, M., {Xu}, B.,
  {Warde-Farley}, D., {Ozair}, S., {Courville}, A., and {Bengio}, Y.
\newblock {Generative Adversarial Networks}.
\newblock \emph{arXiv e-prints}, art. arXiv:1406.2661, Jun 2014.

\bibitem[Hall et~al.(1999)Hall, Brooks, and Doswell~III]{hall_brooks}
Hall, T., Brooks, H., and Doswell~III, C.
\newblock Precipitation forecasting using a neural network.
\newblock \emph{Weather and Forecasting - WEATHER FORECAST}, 14, 06 1999.
\newblock
  \doi{10.1175/1520-0434(1999)014$\langle$0338:PFUANN$\rangle$2.0.CO;2}.

\bibitem[{Ioffe} \& {Szegedy}(2015){Ioffe} and {Szegedy}]{bnorm}
{Ioffe}, S. and {Szegedy}, C.
\newblock {Batch Normalization: Accelerating Deep Network Training by Reducing
  Internal Covariate Shift}.
\newblock \emph{arXiv e-prints}, art. arXiv:1502.03167, Feb 2015.

\bibitem[{Jitkrittum} et~al.(2016){Jitkrittum}, {Szabo}, {Chwialkowski}, and
  {Gretton}]{mean_embed}
{Jitkrittum}, W., {Szabo}, Z., {Chwialkowski}, K., and {Gretton}, A.
\newblock {Interpretable Distribution Features with Maximum Testing Power}.
\newblock \emph{arXiv e-prints}, art. arXiv:1605.06796, May 2016.

\bibitem[{Kingma} \& {Ba}(2014){Kingma} and {Ba}]{kingma2014adam}
{Kingma}, D.~P. and {Ba}, J.
\newblock {Adam: A Method for Stochastic Optimization}.
\newblock \emph{arXiv e-prints}, art. arXiv:1412.6980, Dec 2014.

\bibitem[Koizumi(1999)]{Koizumi}
Koizumi, K.
\newblock An objective method to modify numerical model forecasts with newly
  given weather data using an artificial neural network.
\newblock \emph{Weather and Forecasting - WEATHER FORECAST}, 14:\penalty0
  109--118, 02 1999.
\newblock
  \doi{10.1175/1520-0434(1999)014$\langle$0109:AOMTMN$\rangle$2.0.CO;2}.

\bibitem[Lu \& Ricciuto(2019)Lu and Ricciuto]{gmd-12-1791-2019}
Lu, D. and Ricciuto, D.
\newblock Efficient surrogate modeling methods for large-scale earth system
  models based on machine-learning techniques.
\newblock \emph{Geoscientific Model Development}, 12\penalty0 (5):\penalty0
  1791--1807, 2019.
\newblock \doi{10.5194/gmd-12-1791-2019}.

\bibitem[Rasp et~al.(2018)Rasp, S.~Pritchard, and Gentine]{rasp}
Rasp, S., S.~Pritchard, M., and Gentine, P.
\newblock Deep learning to represent subgrid processes in climate models.
\newblock \emph{Proceedings of the National Academy of Sciences}, 115:\penalty0
  201810286, 09 2018.
\newblock \doi{10.1073/pnas.1810286115}.

\bibitem[Schneider et~al.(2017)Schneider, Lan, Stuart, and Teixeira]{Schneider}
Schneider, T., Lan, S., Stuart, A., and Teixeira, J.
\newblock Earth system modeling 2.0: A blueprint for models that learn from
  observations and targeted high-resolution simulations.
\newblock \emph{Geophysical Research Letters}, 08 2017.
\newblock \doi{10.1002/2017GL076101}.

\bibitem[{Shi} et~al.(2015){Shi}, {Chen}, {Wang}, {Yeung}, {Wong}, and
  {Woo}]{2015arXiv150604214S}
{Shi}, X., {Chen}, Z., {Wang}, H., {Yeung}, D.-Y., {Wong}, W.-k., and {Woo},
  W.-c.
\newblock {Convolutional LSTM Network: A Machine Learning Approach for
  Precipitation Nowcasting}.
\newblock \emph{arXiv e-prints}, art. arXiv:1506.04214, Jun 2015.

\bibitem[{Sutherland} et~al.(2016){Sutherland}, {Tung}, {Strathmann}, {De},
  {Ramdas}, {Smola}, and {Gretton}]{mmd}
{Sutherland}, D.~J., {Tung}, H.-Y., {Strathmann}, H., {De}, S., {Ramdas}, A.,
  {Smola}, A., and {Gretton}, A.
\newblock {Generative Models and Model Criticism via Optimized Maximum Mean
  Discrepancy}.
\newblock \emph{arXiv e-prints}, art. arXiv:1611.04488, Nov 2016.

\bibitem[Taylor et~al.(2011)Taylor, Ronald, and Meehl]{cmip5}
Taylor, K.~E., Ronald, S., and Meehl, G.
\newblock An overview of cmip5 and the experiment design.
\newblock \emph{Bulletin of the American Meteorological Society}, 93:\penalty0
  485--498, 11 2011.
\newblock \doi{10.1175/BAMS-D-11-00094.1}.

\bibitem[{Vondrick} et~al.(2016){Vondrick}, {Pirsiavash}, and
  {Torralba}]{videogan}
{Vondrick}, C., {Pirsiavash}, H., and {Torralba}, A.
\newblock {Generating Videos with Scene Dynamics}.
\newblock \emph{arXiv e-prints}, art. arXiv:1609.02612, Sep 2016.

\bibitem[Watanabe et~al.(2010)Watanabe, Suzuki, O’ishi, Komuro, Watanabe,
  Emori, Takemura, Chikira, Ogura, Sekiguchi, et~al.]{watanabe2010improved}
Watanabe, M., Suzuki, T., O’ishi, R., Komuro, Y., Watanabe, S., Emori, S.,
  Takemura, T., Chikira, M., Ogura, T., Sekiguchi, M., et~al.
\newblock Improved climate simulation by miroc5: mean states, variability, and
  climate sensitivity.
\newblock \emph{Journal of Climate}, 23\penalty0 (23):\penalty0 6312--6335,
  2010.

\bibitem[Weber et~al.(2019)Weber, Corotan, Hutchinson, Kravitz, and Link]{acp}
Weber, T., Corotan, A., Hutchinson, B., Kravitz, B., and Link, R.
\newblock Technical note: Deep learning for creating surrogate models of
  precipitation in earth system models.
\newblock \emph{Atmospheric Chemistry and Physics Discussions}, pp.\  1--16, 04
  2019.
\newblock \doi{10.5194/acp-2019-85}.

\bibitem[{Wu} et~al.(2019){Wu}, {Kashinath}, {Albert}, {Chirila}, {Prabhat},
  and {Xiao}]{2019arXiv190506841W}
{Wu}, J.-L., {Kashinath}, K., {Albert}, A., {Chirila}, D., {Prabhat}, and
  {Xiao}, H.
\newblock {Enforcing Statistical Constraints in Generative Adversarial Networks
  for Modeling Chaotic Dynamical Systems}.
\newblock \emph{arXiv e-prints}, art. arXiv:1905.06841, May 2019.

\bibitem[{Xu} et~al.(2015){Xu}, {Wang}, {Chen}, and {Li}]{leakyrelu}
{Xu}, B., {Wang}, N., {Chen}, T., and {Li}, M.
\newblock {Empirical Evaluation of Rectified Activations in Convolutional
  Network}.
\newblock \emph{arXiv e-prints}, art. arXiv:1505.00853, May 2015.

\end{thebibliography}
\bibliographystyle{icml2019}

\newpage
\appendix
\section*{Appendix A: Evaluation Metrics} 

Evaluating GANs is notoriously difficult.
For our task, we need to show that the joint probability distribution of ESM outputs is the same as joint probability distribution of the generator outputs, which requires a statistic that measures the discrepancy between the two distributions. Of that statistic, we first need to know its statistical distribution when the ESM and generator distributions are the same, which is the ``null-hypothesis distribution'' for the statistic. Second, we need to know how the statistic is distributed in the presence of some de minimus discrepancy (i.e., a discrepancy small enough that even if we knew it was present, we would still be willing to use the model)
between the ESM and generator distributions. We use that to compute the power of our statistical test. If the power of the test is high, and the test fails to reject the null hypothesis, then we can conclude with high confidence that the two distributions are the same. We are exploring Maximum Mean Discrepancy (MMD) \cite{mmd} and Mean Embeddings (ME) \cite{mean_embed} metrics for evaluation of the model.

\section*{Appendix B: Dataset Details} \label{app:data}

\begin{table}[ht]
  \caption{MIROC5 data for the model}
  \label{tab:data}
  \centering
  \begin{tabular}{lll}
    \toprule
     Scenario & Realizations & Years \\
    \midrule
    Historical & r1i1p1-r5i1p1 & 1950-2009\\
    RCP\{2.6, 4.5, 6.0, 8.5\} & r1i1p1-r3i1p1 & 2006-2100\\
    RCP\{2.6, 4.5, 6.0, 8.5\} & r4i1p1-r5i1p1 & 2006-2035\\
    \bottomrule
  \end{tabular}
\end{table}

\end{document}